# Replicating a Disjoint-Set Union Experiment over Various Notions of Micro Units to assess Translation Effort

Michael Carl, CRITT, Kent State University

Miljanović et al (2025) present a study to assess "the distribution of editing procedures across micro and macro units as an indicator of the strain of text production" (p. 12). They segment keystroke logging data into sequences of temporal 'micro units' (MUs) based on the median inter-keystroke (IKI) duration between two alpha-numeric keystrokes (i.e., within-word) plus 2 STD, which we refer to here as MUD (cf. Bandaru et al 2026). An MU is thus a stretch of continuous text production based on the typist's (i.e., translator's) average typing speed. The stretch of text covered by an MU can be further revised during successive MUs in a drafting and/or a revision phase, leading to overlapping textual segmentations. The set of overlapping MUs, i.e., an initial MU plus all successive revisions form a temporal macro unit (MX).

Miljanović et al adopt (Alves & Vale 2011, Alves & Gonçalves 2013) approach to distinguish between four types of MX:

- P0: the MX consists of only one MU during initial text production.
- P1: the MX has one or more additional MUs only during the drafting phase.
- P2: the MX has one or more additional MUs only during end revision.
- P3: the MX consists of MUs during drafting and end revision.

They describe a manual annotation process by which "all micro units belonging to one macro unit are connected" and labeled. In formal terms, an MX seems to be the transitive closures over MUs, i.e., the set of all MUs that are (directly or indirectly) linked together through sharing one or more common word(s). The transitive closure results in a disjoint-set union (DSU), a set of sets (i.e., a set of MX) where no two MX share any common elements (i.e., IDs of target words).

| # | MU | MX | Type of MU / MX and Comment |
|---|---|---|---|
| $Unit_1$ | 1+2 | 1+2 | P0: $MU_1$ keystrokes related to word IDs 1+2 |
| $Unit_2$ | 2+3 | 1+2+3 | P1: $MU_2$ shares word Id 2 with $MU_1$ , leading to $MX_2$ |
| $Unit_3$ | 4+5 | 4+5 | P0: $MU_3$ keystrokes covering word IDs 4+5 |
| $Unit_4$ | 5+6 | 4+5+6 | P1: $MU_4$ shares word Id 5 with $MU_3$ , leading to $MX_4$ |
| $Unit_5$ | 3+4 | 1+2+3+4+5+6 | P3: $MU_5$ bridges $MX_2$ and $MX_4$ , leading to $MX_5$ |

*Table 1: Example of a disjoint-set union of Target Text Indexes (TTid) grouped into MUs and MX*

To illustrate this notion, Table 1 shows the sequence of 5 MUs in which a (hypothetical) text of six words is produced (target token indexes TTid 1 to 6). Assume, the first four MUs are part of a drafting phase while $Unit_5$ takes place during the revision phase. The numbers in the MU and MX columns indicate word IDs. Whereas $MU_1$ and $MU_3$ are of type P0, the overlap of keystrokes related to word 2 in $MU_2$ and to word 5 in $MU_4$ results in the formation of $MX_2$ and $MX_4$ respectively. $MU_5$ bridges $MX_2$ and $MX_4$ which leads to the construction of $MX_5$.

As Table 1 demonstrates, the DSU in this example consists of one large MX of type P3 (containing MUs during drafting and revision) which contains all six target word IDs, thus subsuming all other MUs. Without the revision $MU_5$, the DSU of this session would contain two macro units, $MX_2$ and $MX_4$.

In their experiment with 38 participants, Miljanović et al collected 152 translation sessions, each with approximately 250 words. This can, theoretically, yield a DSU with between a total of 152 MX - one for each text – and a maximum of 152*250 = 38000 MX, one MX for each word. Based on their manual MX-annotation procedure they observe a total of 3096 MX in the 152 sessions, between 11 and 25 MX per session. They find that "P0 is overall the most frequent category" while "P2 and P3 editing strategies were overall very infrequent." However, they do not reveal how many words each MX entail, nor how many MUs each MX contains.

**A replication experiment**

Since the shape of MUs, and thus the set of MXs, depends on the pause threshold fragmenting the flow of keystrokes into MUs, we replicate Miljanović et al based on CRITT TPR-DB data (Carl et al 2016), comparing five different segmentation methods. We aim to investigate the impact of the threshold (i.e., the segmentation method) on the editing strategies, as defined by Miljanović et al.

We use the CRITT TPR-DB 3.0 which has a fully automatic keystroke-to-word mapping algorithm[1] which associates every keystroke (insertions and deletions) with a target word ID (TTid). We then automatically assign sets of TTids to MUs in the following way:

1) We extracted from the TPR-DB[2] a set of 478 from-scratch translation sessions, involving eight different language pairs with a total of 70302 target words.
2) We compute MUs based on five different segmentation methods:
    a. TSP, MUD and PUB are based on translator specific segmentation thresholds, as discussed in Bandaru et al (2026, this conference)
    b. K1000 is a static threshold of 1000ms (Carl 2016) that has previously been used in the CRITT TPR-DB to separate production units (PUs).
    c. A TT-based MU, defined as a coherent sequence of keystrokes associated with the same TTid; the IKI (keystroke pause) preceding a TT-based MU is

[1] This algorithm is described on the TPR-DB 3.0 documentation site: https://critt-kent.github.io/TPR-DB-documentation/process/automatic-processing/#keystroke-to-word-mapping. An earlier version of keystroke-to-word mapping used in TPR-DB 2 was described in (Carl 2012)
[2] Extracted from public studies see https://sites.google.com/site/centretranslationinnovation/tpr-db/public-studies

the IKI between two word. A TT-based MX aggregates all MUs with the same TTid. (see Carl 2024)

3) We implemented a Python script that automatically extracts and labels the DSU for each of the 478 sessions, thus avoiding manual annotation.[3]

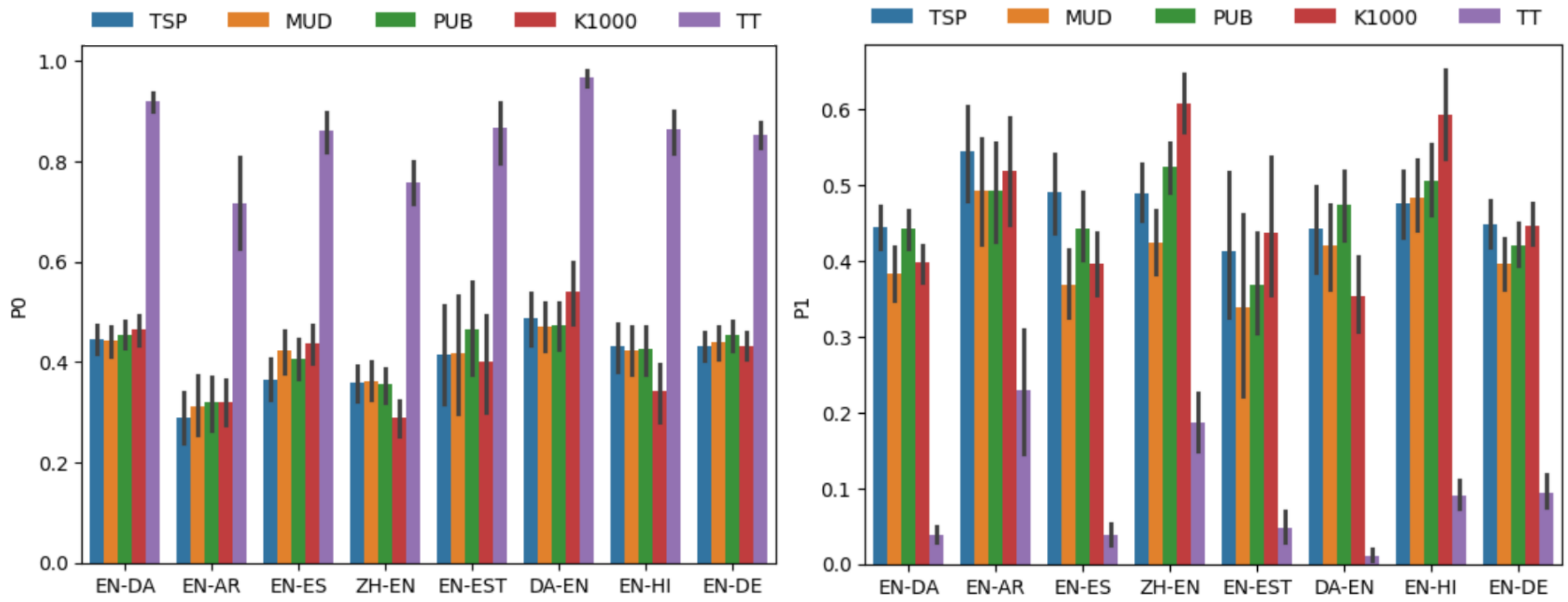

*Figure 1: Percentage of P0 (left) and P1 for the different Segmentation methods and language pair: English (EN) to Arabic (AR), Spanish (ES), Estonian (EST), Hindi(HI) and German (DE), and Chinese (ZH) to English, Danish (DA) to English..*

| Seg-Type | P0 | P1 | P2 | P3 | P4 | P{1,2,3,4} |
|---|---|---|---|---|---|---|
| **K1000** | 40.71 | 46.78 | 1.49 | 5.61 | 4.14 | 58.02 |
| **MUD** | 41.51 | 40.99 | 2.52 | 8.73 | 4.95 | 57.19 |
| **PUB** | 42.05 | 46.28 | 1.48 | 6.22 | 3.97 | 57.95 |
| **TSP** | 40.75 | 46.82 | 1.34 | 6.01 | 4.19 | 58.36 |
| **TT** | 85.42 | 9.30 | 1.86 | 0.38 | 3.04 | 14.58 |

*Table 2: Percentage of translation strategies as detected per segmentation method.*

## Evaluation

For the 478 translation sessions, grouped into eight different language pairs, the graph on the left side in Figure 1 shows the percentage of MX with P0-strategy, the graph on the right side the distribution of MX under P1-strategies. As can be seen the percentage of the P0 strategy is lower for English-to-Arabic (EN-AR) and for Chinese-to-English (ZH-EN) as compared to the other language pairs, but tends to be higher for P1 strategy. The four time-based segmentation methods (TSP, MUD, PUB, K1000) provide approximately similar distributions, which are, however, very different from the TT-based segmentation

[3] The script MicroUnits.ipynb can be downloaded from https://github.com/Critt-Kent/Key-Gaze-Correlation.git

method.Table 2 shows the overall frequencies of the editing strategies: According to this table, P0 and P1 cover approximately a similar share of strategies, and their sum amounts to between 82% and 87%. Thus, our data does not confirm Miljanović et al. in that P0 is always more frequent than P1, but our findings support that "P2 and P3 editing strategies were overall very infrequent." However, in addition to the four strategies listed by Miljanović et al., we also found in our data instances of P4, that is, translators edited a word for the first time during the revision phase. Interestingly, the P4 strategy is more frequent than P2 in our data. Very different from the four time-based segmentation methods, Table 2 (and Figure 1) shows that the TTid-based segmentation accumulates roughly 85% of editing strategies under P0 and less than 10% under P1 strategy. However, similar to the other segmentation methods, P0 seems to be relatively less frequent and P1 relatively more frequent for EN-AR and EN-ZH as compared to the other six language pairs.

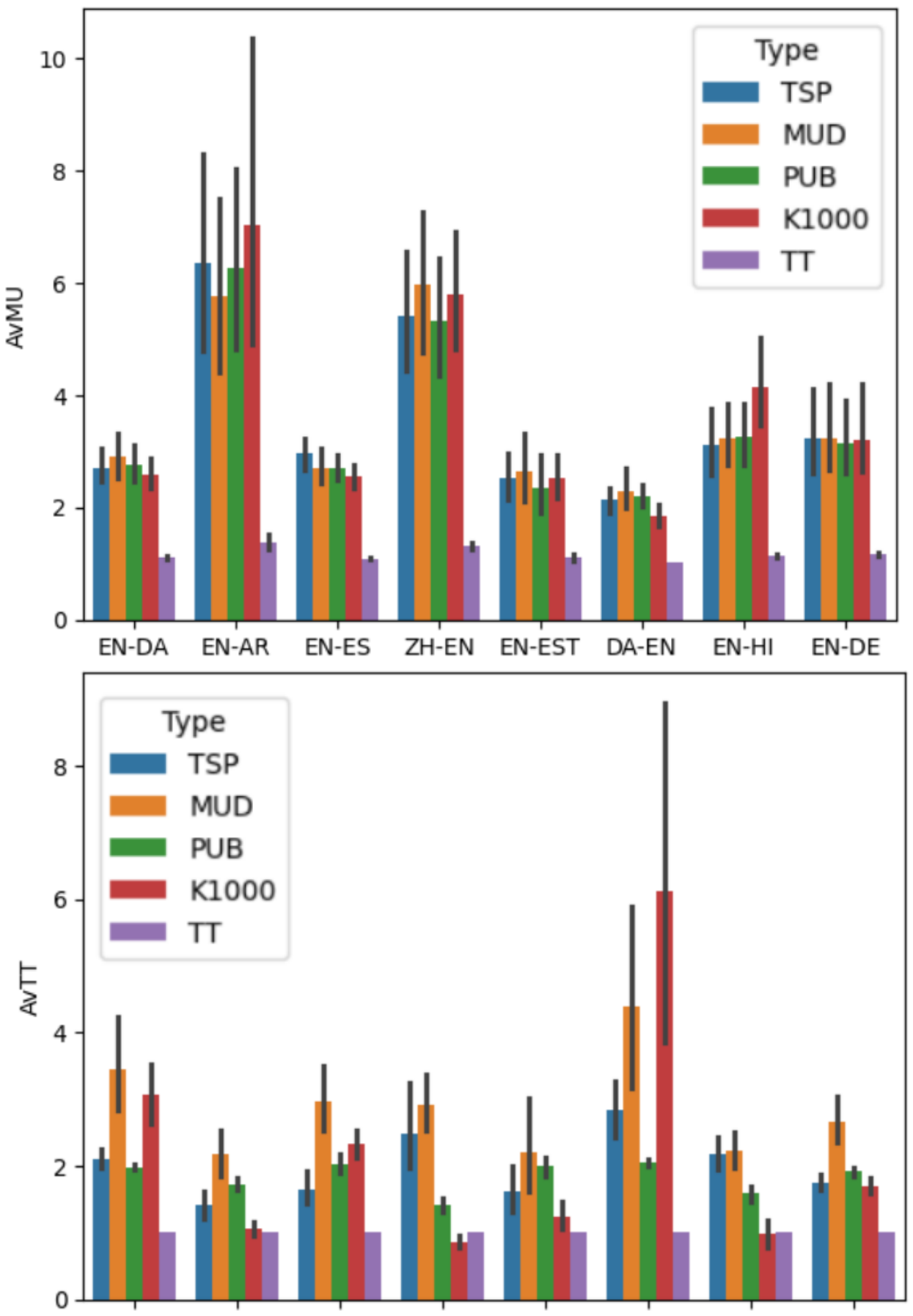

*Figure 2: Average number of MUs per MUX (AvMU, top) and average number of TT words per MUX (AvTT, bottom) in relation to language pair (English to Arabic, Spanish, Estonian, Hindi and German, as well as Chinese to English and Danish to English) and type of segmentation threshold.*

| | K1000 | MUD | PUB | TSP | TT |
|---|---|---|---|---|---|
| **TTid** | 2.2 | 3 | 1.82 | 2.06 | 1 |
| **MU** | 3.7 | 3.7 | 3.55 | 3.59 | 1.17 |

*Table 3: Average number of TTid and MUs per MX and segmentation method.*

Figure 2 plots the relation between the number of MUs per MX (top), and the number of TTids covered per MX (bottom). The average number of MUs (AvMU) per MX is significantly different for EN-AR and ZH-EN as compared to the other language directions. While there are, on average, around 6 MUs per MX for EN-AR and ZH-EN there are 3 to 4 MUs for EN-HI and EN-DE and even fewer for the other language pairs. Note that this relation is inverse proportional to the percentage of P0 in Figure 1 (left): the P0 strategy implies one MU per MX, see Figure 2 (top).

The consistently lowest number of MUs is, however, observed for TTid-based segmentation, where an MU is defined as a coherent sequence of keystrokes related to the same word (TTid), irrespectively of the word-internal IKI duration. Thus, if, during the production of one word, a typist pauses longer than the segmentation-specific threshold, this counts as two MUs for the time-based segmentation methods. For the TTid method, however, all successive keystrokes are grouped into the same MU irrespective of the duration of their IKI. This effect can also be observed in the distribution of AvTT in the bottom part in Figure 2: for the TT method, each MX covers, by definition, only one TTid (see Table 3). For the time-based segmentation methods an MX covers in most cases 2 or more TTids. As shown in Table 3, a MUD-based MX covers on average 3 TTids and 3.77 MUs, whereas in the TT method each MX corresponds to exactly 1 TTid with an average of 1.17 MUs.

## Conclusion

As this investigation shows, each pause threshold fragments the keystroke data in different ways which may lead to different properties of micro and macro units. Our analysis shows a stark difference between the temporal segmentation methods (TPD, MUD, PUB, K1000) a text-based method (TT). The former methods show great similarity, for instance in the number of MUs that each MX covers (Table 3) and the distribution of P0 and P1 translation strategies, yet they systematically differ from the text-based TT approach. However, temporal segmentation methods risk obscuring linguistically meaningful revision behavior by fragmenting production according to pauses rather than units of textual structure. The static K1000 and the translator-specific MUD segmentation methods may blur the precision of the conclusions, given the large(r) number of words covered by each MX (TTid, see Table 3), depending on the research question, might turn out to be less reliable.

However, temporal segmentation methods risk obscuring linguistically meaningful revision behaviour by fragmenting production according to pauses rather than units of textual structure. As a method covers a larger number of words per MX unit, such as the static K1000 and the translator-specific MUD (see Table 3), the analytical precision reduces, in particularly for research questions where the exact scope of a revision target matters.

This may become problematic when investigating phenomena such as cohesive chains, where the relevant unit of analysis is inherently linguistic rather than temporal. From this perspective, the use of temporal MUs to assess DSU properties in relation to specific linguistic patterns appears theoretically misaligned. The question of whether a pause within the production of a single word should constitute a separate MU highlights a deeper conceptual issue: temporal segmentation captures fluctuations in typing behavior, but not necessarily revisions at the level of meaning or textual cohesion. Consequently, it may introduce distortions, such as inflated counts of revision-related units (e.g., P1), without clear interpretive value for the phenomenon under study.

By contrast, the TT segmentation, in which each MX corresponds to exactly one target-text word, offers a linguistically grounded and analytically more transparent framework. Thus, the notion of product-based MUs provides a direct way of capturing revision behavior: initial production (P0) and subsequent modifications (P1, P2, P3) are anchored to stable textual units rather than inferred from temporal thresholds.

The CRITT TPR-DB provides readily available features, which allow researchers to investigate revision dynamics in a way that is both methodologically consistent and theoretically aligned with the object of inquiry—rendering the computation of DSUs unnecessary for this purpose.

## References

Alves, Fabio & Gonçalves, José L. 2013. Investigating the conceptual-procedural distinction in the translation process. A relevance-theoretic analysis of micro and macro translation units. Target 25(1): 107–24.

Alves, Fabio & Vale, Daniel. 2011. On drafting and revision in translation. A corpus linguistics-oriented analysis of translation process data. TC3: Translation: Corpora, Computation and Cognition 1(1): 105–122.

Bandaru, Devi Sri; Michael Carl and Xinyue REN (2026) Assessing Pause Thresholds for empirical Translation Process Research. Translation in Transition 8, (

Carl, Michael & Schaeffer, Moritz & Bangalore, Srinivas. 2016. The CRITT translation process research database. In Carl, Michael & Bangalore, Srinivas & Schaeffer, Moritz (eds.), New Directions in Empirical

Carl, Michael. 2012. Translog-II: A program for recording user activity data for empirical reading and writing research. In Proceedings of the Eight International Conference on Language Resources and Evaluation (LREC'12). (http://www.lrecconf.org/proceedings/lrec2012/pdf/614_Paper.pdf)

Carl, Michael. 2021. Micro units and the first translational response universal. In Explorations in Empirical Translation Process Research. Cham: Springer. 233–57. doi: 10.1007/978-3-030-69777-8_9.

Miljanović, Zoë & Fabio Alves & Celina Brost, & Stella Neumann. 2025. Directionality in translation: Throwing new light on an old question. SKASE Journal of Translation and Interpretation, 2025; 18(2): 4–37. doi: 10.33542/JTI2025-S-2